# A Multi-Agent LLM Framework for Personalized Health Checkup Interpretation and Guidance

HyungJun Kim[1†], Taehan Lee[1†], Soojin Cheon[1†]

[1]DaNaA Data Co., Ltd.

hjkim1568@danai.co.kr; rendezvous.ted@danai.co.kr; soooz819@danai.co.kr

† These authors contributed equally to this work.

**Abstract**

Personalized interpretation of health checkup results requires reasoning across longitudinal records, medical knowledge, lifestyle guidance, and healthcare navigation. We present a multi-agent large language model (LLM) system that identifies multiple intents, maps each to a task-specific agent, executes them in parallel, and synthesizes their outputs. We compared answers generated in Single Agent and Multi Agent settings on 120 Korean compound queries combining two to four requirements, using synthetic health checkup records. The Multi Agent improved the weighted LLM-judge score from 1.695 to 1.797 ($p = 0.027$), and three additional LLM judges showed consistent improvements ($\Delta = +0.111$ to $+0.186$, all $p < 0.05$). The gains came from usefulness, consistency, and the handling of every requirement in compound queries, whereas numerical accuracy and grounding improved significantly under only one of the four judges and medical safety did not differ, and critical failures occurred at similar rates (Single Agent 15.0% vs. Multi Agent 13.3%). Two human evaluators preferred Multi Agent in 66.7% and 68.3% of pairwise comparisons. Multi Agent execution increased latency and cost by 1.31× and 2.02×, respectively. In exploratory subgroup analyses, the improvement was concentrated in queries involving personal-record lookup.

# 1. Introduction

Korea's national general health screening program operates at population scale [1, 2], but its reports mainly present test values with labels such as "normal" or "abnormal." They provide limited support for interpreting longitudinal changes, assessing clinical relevance, and identifying appropriate follow-up actions. LLM-based medical systems may help address this gap [3–5], but health-checkup counseling extends beyond general medical question answering. It may require personal-record retrieval, use of medical references, interpretation of user-provided measurements, lifestyle guidance, or healthcare service information.

A single query can combine several of these tasks. For example, a user may ask the system to review a blood pressure trend, assess whether it is concerning, and recommend lifestyle changes. Answering the full query requires coordinated use of personal records, medical evidence, individualized judgment, and recommendations. A single-agent system with a unified prompt may make it difficult to separate task-specific evidence and apply task-specific safeguards, particularly in long contexts [6]. This is especially important in healthcare settings, where unsupported medical claims can be harmful [7, 8]. In our study, compound queries sometimes received a detailed answer to one requirement and little or no response to another. This failure motivated explicit query decomposition rather than reliance on a single undifferentiated reasoning process.

We therefore developed an intent-driven framework that maps each query to one or more task-specific intents, routes the identified intents to task-specific agents, and executes multiple agents concurrently when applicable. A synthesis module then combines their outputs into a coherent response. This study evaluates whether executing multiple agents in this way improves responses to compound health-checkup queries and how much latency and cost it adds. Unlike systems in which agents discuss or collaborate, such as MedAgents [9], AutoGen [10], and MetaGPT [11], our agents do not deliberate with one another; they execute predefined task-specific roles selected by an explicit intent taxonomy and deterministic intent-to-agent mapping.

This work contributes (1) an intent-based orchestration framework for parallel execution of heterogeneous health-checkup tasks, including explicit handling of compound queries; (2) a routing layer with stage-wise model assignment and tracing for auditable routing decisions; and (3) an evaluation on a labeled Korean benchmark covering answer quality, critical failures, latency, and cost.

# 2. Related Work

## 2.1. LLM-Based Multi-Agent Systems

Most LLM-based multi-agent systems address complex tasks through collaboration or interaction among multiple agents. CAMEL uses structured role-playing dialogue [12]; AutoGen composes agents, humans, and tools [10]; and MetaGPT systematizes collaboration through standardized workflows [11]. Survey studies further examine communication structures and capability acquisition across such systems [13]. Medical systems follow similar collaborative patterns: MedAgents connects expert agents that discuss clinical questions iteratively until consensus [9], whereas MDAgents selects solo or group reasoning according to query difficulty [14]. In contrast, our framework decomposes a compound counseling query into predefined task-specific roles, executes the corresponding agents in parallel, and uses a synthesizer to combine their outputs.

## 2.2. Orchestration and Model Routing

Related routing work generally optimizes model cost and quality. FrugalGPT escalates queries through stronger models when a cheaper answer fails a learned check [15], while RouteLLM learns whether to use a strong or weak model from preference data [16]. HuggingGPT instead uses an LLM controller to decompose a request into subtasks and dispatch them to expert models [17]. Our router operates on task intent: a classifier emits intents, and a deterministic mapping selects the corresponding agents. Model selection occurs separately through configured assignments for classification, context extraction, sub-agents, and synthesis. This separation makes routing decisions auditable at the intent level.

## 2.3. LLMs for Healthcare and Personalized Health Guidance

Medical LLM studies have established strong question-answering performance [3, 4] and favorable ratings on patient questions [5]. HuatuoGPT targets Chinese medical consultation [18], while PH-LLM grounds sleep and fitness coaching in longitudinal wearable data [19]. Evaluation on Korean medical licensing examinations suggests that performance on English benchmarks may not directly generalize to Korean medical contexts [20]. Hallucination remains a concern for LLMs in medical settings [7], and even recent models struggle to detect medical hallucinations reliably [8].

Although health checkups are widely used at the population level [1, 2], research on personalized interpretation and guidance based on individuals' longitudinal screening records remains relatively limited. Such applications require not only strong question-answering performance but also reliable grounding in individual records and effective decomposition of complex user queries into appropriate tasks. To address these requirements, our system grounds responses in individual screening records and domain-specific retrieval stores [21], and routes identified intents to task-specific agents.

# 3. Method

## 3.1. Problem Formulation

At turn $t$, the system receives a user query $u_t$, dialogue history $H_t$, and a multi-year health checkup record $D$. The intent taxonomy $I$ contains 29 intent codes grouped into seven intent-bearing agent categories plus a follow-up-suggestion category, and the slot schema $S$ contains 13 categories covering entities such as test and disease names, symptoms, dates, health-status trends, hospitals and regions, health concerns, and user-supplied values. Given ($u_t$, $H_t$), the Multi-Agent Orchestrator predicts a set of intents and associated slots, along with candidate agents. Each intent is mapped to a task-specific agent, whose outputs are subsequently combined to generate the final response $y_t$ and suggested follow-up questions.

## 3.2. System Architecture

Figure 1 illustrates the multi-agent system pipeline. A rule-based pre-filter rejects queries containing predefined personal identifiers or prompt-injection patterns. Intent classification and RAG-based document retrieval run concurrently, after which deterministic rules normalize slots and resolve follow-up queries. A lightweight model selects relevant sections from retrieved health records and reference documents, while the memory system supplies recent turns, long-term facts, a recap, and tone preferences. The selected agents then execute concurrently, and a synthesizer integrates their outputs into the final response.

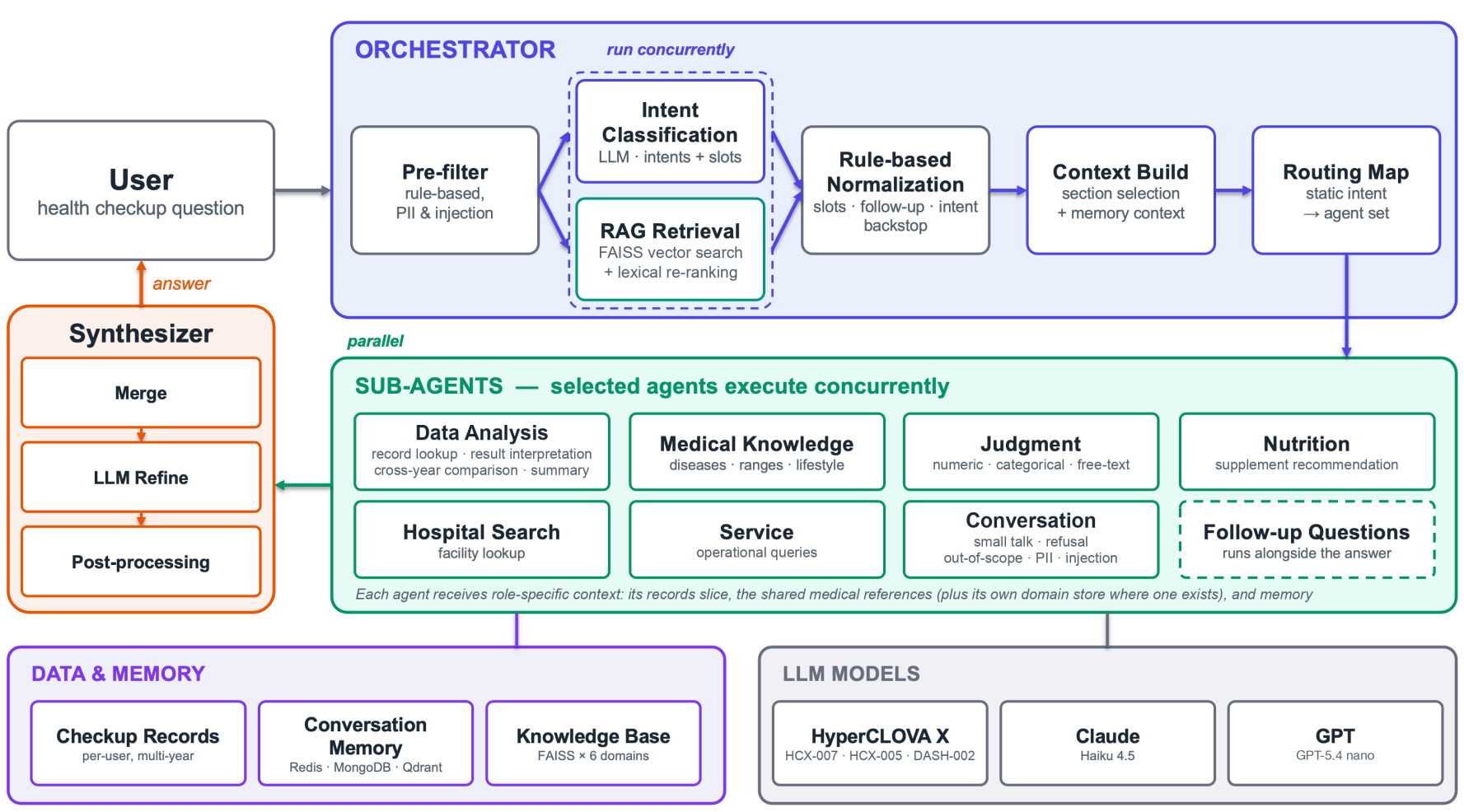

**Figure 1. System architecture.** Predicted intents are deterministically mapped to task-specific agents (Table A1), which receive role-specific context from screening records, retrieved documents, and memory. Applicable agents run concurrently, and the synthesizer combines their outputs.

## 3.3. Orchestrator and Intent Routing

The Multi-Agent Orchestrator returns intents, slots, and candidate agent names in a single call. It validates candidates against the predefined routing map, assigns mapped agents when needed, removes unknown or duplicate intents, propagates shared slots, and limits the intent set to four. Single-agent-only intents, namely the three refusal intents and hospital search, take precedence over other intents. Deterministic normalization resolves relative dates, whole-history requests, and years beyond the latest available screening.

Two rule-based stages then refine the prediction without additional model calls. A backstop stage adds an omitted intent when explicit query cues support it and corrects the judgment type when necessary, without removing predicted intents. A specialization stage replaces generic medical-knowledge codes with more specific codes, without changing the executed agent set.

Dialogue rules address cases where context causes self-contained queries to be misinterpreted as follow-ups. They inherit or transform slots from the preceding turn for elliptical queries, while self-contained compound queries are classified independently of dialogue history. These rules ensure that queries requiring multiple agent roles retain the relevant intents and execute the corresponding agents.

## 3.4. Specialized Sub-Agents

Table A1 lists 29 intent codes across seven intent-bearing categories; together with the follow-up suggestion category, the eight categories comprise 14 agents, 13 of which are intent-mapped. Data analysis handles record lookup, result interpretation, cross-year comparison, and whole-report summarization using checkup records as the primary personal health source. Medical knowledge addresses general questions on diseases, reference ranges, and lifestyle, combining relevant record values with domain knowledge when needed. Three judgment agents evaluate user-supplied findings as numeric, categorical, or free-text inputs. Nutrition provides supplement recommendations based on stated concerns, while hospital search and service handle region-specific facility and operational queries. Conversation includes a default small-talk agent and a single refusal agent covering all three refusal intents: personal-data and prompt-injection or system-probe queries passing the pre-filter receive the same fixed response as the pre-filter without a model call, whereas out-of-scope requests such as product or medication recommendations receive a model-generated refusal. A separate follow-up suggestion agent concurrently generates three follow-up questions.

Six categories maintain separate FAISS vector stores [22] built from curated domain documents for retrieval-augmented generation [21]; the conversation and follow-up suggestion categories do not retrieve documents. All vector-store documents were reviewed by a clinician at a tertiary hospital. Vector-search candidates sharing no terms with the query are down-weighted by lexical re-ranking. A shared medical-reference retrieval runs once per turn and is available to all executing agents, while agents with domain-specific stores additionally query them. For personal records, a lightweight model selects up to eight relevant test-item codes, which are then used for exact-code filtering of the multi-year record. Before generation, a grounding check verifies the retrieved context. Unavailable years or topics are reported as such rather than inferred.

## 3.5. Response Synthesis and Hierarchical Memory

Response synthesis consists of three stages. First, when multiple agents are executed, their outputs are merged in a fixed order: personal record values, general explanations, and recommendations; when a judgment agent runs with a record agent, the general block containing the judgment is placed first.

Second, an LLM refines the merged blocks, or the single-agent output, together with the user query. It preserves every requirement addressed in the agent outputs in the order asked, removes redundancy, resolves conflicting record values in favor of the screening record, and applies safety rules, which take precedence over this preservation, to remove or soften unsupported medical recommendations, drug information, definitive efficacy claims, and statements inconsistent with the recorded status. Personal record values are passed separately with instructions to preserve cited values, units, verdicts, and years; single-year results are quoted as is, whereas series spanning three or more years are condensed to the latest value and a trend sentence unless the query asks for a comparison or trend. Finally, rule-based post-processing removes LaTeX formatting and meta-commentary, softens overly definitive medical statements, and removes sentences whose year-attributed values do not match the screening record.

Memory is organized hierarchically: Redis and MongoDB hold recent dialogue turns, Qdrant stores embedded health facts and recurring concerns, and a user profile keeps conversation recaps and tone preferences. An extractor and a decider update these stores asynchronously, and relevant long-term items are retrieved at response time by vector similarity and slot matching. The intent classifier receives only the recent dialogue turns; long-term items are used at response generation.

## 3.6. Stage-Wise Model Assignment and Tracing

Each pipeline stage uses its assigned model: lightweight models for classification, extraction, question suggestion, and refinement; reasoning models for data analysis and medical knowledge; and other models for judgment, nutrition, service, and hospital search (Section 4.1). HyperCLOVA X calls are retried with an alternative model when rate-limited. A centralized tracing layer records the provider, model, calling agent, token counts, and latency for all model calls. Retrieval and other non-model operations are logged as tool events under the same turn-level trace ID. These traces provide the metrics reported in Section 4.

# 4. Experiments

## 4.1. Experimental Setup

We evaluated the system using a Korean-language benchmark and turn-level execution traces from the same orchestration pipeline. The benchmark consists of 120 compound queries drawn from four of the seven intent-bearing categories in Table A1: data analysis (record lookup, result interpretation, cross-year comparison, and whole-report summary), medical knowledge, judgment (numeric, categorical, and free-text judgment of user-supplied values), and nutrition (supplement recommendation). The hospital-search, service, and conversation categories are not included: hospital search and the three refusal intents are configured to run alone, and no query targets the remaining service or small-talk intents. Evaluation items were annotated with target intents and, where applicable, ground-truth values derived from predefined health records. All evaluations used five synthetic health-checkup records constructed for this study. Test values were generated to fall into predefined verdict categories (normal, caution, abnormal, and others) under clinically reviewed in-house criteria, and each record contained up to six annual examinations from 2020 to 2025, including abnormal findings and overall-opinion comments.

The Multi Agent configuration uses HyperCLOVA X HCX-007 for the data-analysis and medical-knowledge agents, HCX-005 for the judgment and nutrition agents, Claude-Haiku-4.5 for intent classification and context extraction, and GPT-5.4-nano for test-item extraction and response synthesis; the service and hospital-search agents (HCX-005 and HCX-DASH-002) were outside the evaluation.

Single Agent used the same model and prompt settings as Multi Agent. The orchestrator identified a single intent, selected and executed the corresponding agent, and passed its output through the same synthesizer, including LLM refinement, safety rules, and rule-based post-processing (Section 3.5); only multi-agent output merging was omitted. The pre-filter, memory, and follow-up suggestion modules are described for completeness. The follow-up suggestion agent was not executed and was excluded from call-count and cost calculations, while the pre-filter and memory modules were not evaluated.

## 4.2. Evaluation and Metrics

We compare two configurations. The Multi Agent configuration performs multi-intent detection, agent routing, and parallel execution, allowing multiple agents to be executed for a single query, after which the synthesizer merges their outputs. Single Agent assigns one intent to each query and executes only the corresponding agent, using that agent's prompt and retrieval stores. This answer is then processed by the same synthesizer without the multi-agent merging stage. The refinement step reorganizes and filters the available agent outputs but does not generate content for requirements not addressed by an executed agent. The primary differences between the configurations are whether a single or multiple intents are identified, the resulting number of executed agents, and whether multiple agent outputs are merged; all other settings, including prompts, RAG, and the synthesizer's refinement, safety rules, and post-processing, remain identical.

Routing performance is evaluated using micro-averaged precision, recall, and F1-score, along with exact-set match for the executed agent sets. Efficiency is evaluated using total latency, time to first token, input and output token usage, number of model calls, and cost. Answer quality is assessed using blinded LLM judges [23, 24] with the zero-to-two weighted rubric and critical-failure gate described in Appendices B and C. Multiple judges, including models from providers not used in the generation pipeline, are used to assess whether the results depend on the judge model.

## 4.3. Routing Analysis

Table 1 reports executed-agent coverage for Single Agent and the Multi-Agent Orchestrator, measured against the labeled agent sets. The 120 queries require 277 labeled agent executions (2.31 distinct agents per query). For Single Agent, the executed agent is the agent corresponding to its single assigned intent. Single Agent executed one agent per turn and therefore covered only about one-third of the required agents (micro-averaged precision 0.767, recall 0.332, and F1-score 0.463), with an exact-set match of 0.8% (1/120). The Multi-Agent Orchestrator executed 2.33 agents per turn and achieved a micro-averaged precision of 0.979, recall of 0.989, and F1-score of 0.984, and an exact-set match of 94.2% (113/120). These metrics measure routing fidelity rather than response quality.

**Table 1. Executed-agent coverage of the Single Agent and Multi-Agent Orchestrator on the test set.** All metrics are micro-averaged with respect to the labeled agent sets. For the Single Agent, the executed

agent is defined as the agent corresponding to the assigned intent. Its exact-set match is therefore at most 1/120, as only one query of the test set requires exactly one agent for a single intent.

| Configuration | n | Micro-averaged precision | Micro-averaged recall | Micro-averaged F1-score | Agents per turn | Exact-set match |
|---|---|---|---|---|---|---|
| Single Agent | 120 | 0.767 | 0.332 | 0.463 | 1.00 | 1/120 (0.8%) |
| Multi Agent | 120 | 0.979 | 0.989 | 0.984 | 2.33 | 113/120 (94.2%) |

## 4.4. Answer Quality

Answers were anonymized and scored by blinded judges on seven modules (0–2), combined into a weighted total, with critical failures recorded separately (Appendices B and C). Four LLM judges scored the entire test set. Because Claude-Opus-5.5 and GPT-6-Sol are from the model families used in answer generation, Gemini-3.8-Flash and Solar-Pro-4, from providers not used in the generation pipeline, were designated in advance as independent judges, with Gemini-3.8-Flash as the primary judge.

Table 2 summarizes the answer-quality results. In the Gemini-3.8-Flash evaluation, Multi Agent achieved a higher weighted total score than Single Agent (Single Agent 1.695 ± 0.44 vs. Multi Agent 1.797 ± 0.36, $\Delta = +0.103$, 95% CI [0.012, 0.193], $p = 0.027$), and the proportion of responses with at least one critical failure was similar (Single Agent 18/120 [15.0%] vs. Multi Agent 16/120 [13.3%]). The other three judges showed larger improvements ($\Delta = +0.111$, +0.142, and +0.186; $p = 0.017$, 0.006, and $< 0.001$), making the primary estimate the most conservative.

**Table 2. Weighted answer score by judge model for the test set.** Δ denotes Multi Agent minus Single Agent; p-values are from paired t-tests. Gemini-3.8-Flash is the primary judge.

| Judge model | n | Single Agent | Multi Agent | Δ (95% CI) | p |
|---|---|---|---|---|---|
| Claude-Opus-5.5 | 120 | 1.575 ± 0.42 | 1.686 ± 0.38 | +0.111 (0.020, 0.202) | **0.017** |
| GPT-6-Sol | 120 | 1.460 ± 0.47 | 1.602 ± 0.43 | +0.142 (0.041, 0.244) | **0.006** |
| Gemini-3.8-Flash | 120 | 1.695 ± 0.44 | 1.797 ± 0.36 | +0.103 (0.012, 0.193) | **0.027** |
| Solar-Pro-4 | 120 | 1.613 ± 0.48 | 1.799 ± 0.33 | +0.186 (0.090, 0.281) | **< 0.001** |

Across the seven rubric modules, usefulness, compound-query handling, and consistency improved significantly under all four judges (Figure 2). For the primary judge, usefulness showed the largest improvement ($\Delta = +0.265$), followed by compound-query handling (+0.192) and consistency (+0.133). Computational correctness and medical safety showed no significant difference under any of the four judges, and numerical accuracy and evidential grounding improved only under Solar-Pro-4 ($\Delta = +0.210$ and +0.218; $p = 0.011$ and 0.005). Full per-module results are provided in Table C1.

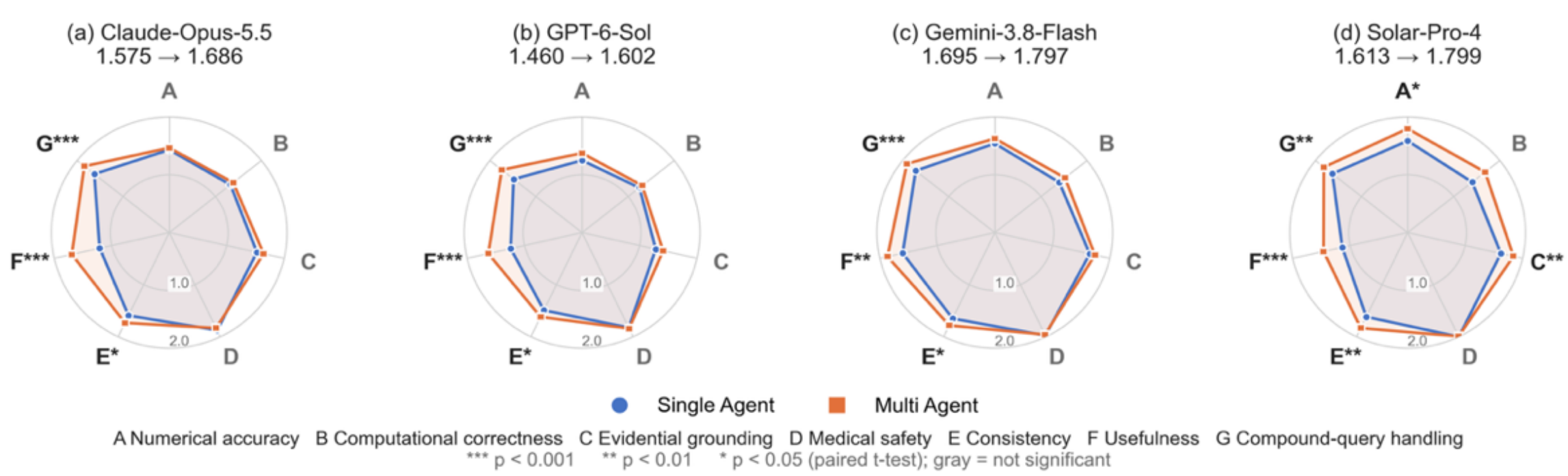

**Figure 2. Rubric module scores by judge model.** Mean scores (0–2) for Single Agent and Multi Agent across 120 paired queries: (a) Claude-Opus-5.5, (b) GPT-6-Sol, (c) Gemini-3.8-Flash, and (d) Solar-Pro-4. Each module is averaged over the pairs in which the judge scored both answers on it; see Table C1 for n, differences, and p-values.

Item-level weighted totals from Claude-Opus-5.5 and Gemini-3.8-Flash correlated at r = 0.878, and module-level agreement ranged from Cohen's κ = 0.464 to 0.832 (Table 3). Because Claude-Haiku-4.5 (intent classification and context extraction) and GPT-5.4-nano (test-item extraction and refinement) are used identically in both configurations, any same-provider bias of Claude-Opus-5.5 or GPT-6-Sol would affect both answers and is therefore unlikely to account for the between-configuration difference. Consistent with this, the two judges from providers not used in the generation pipeline (Gemini-3.8-Flash and Solar-Pro-4) also showed between-configuration differences (+0.103 and +0.186). However, because the synthesizer merges multiple agent outputs only in the Multi Agent configuration, the refinement model processes more content in Multi Agent responses, and residual asymmetry cannot be fully excluded. The lowest agreement values were observed for usefulness (κ = 0.464) and medical safety (κ = 0.472). The latter was accompanied by high exact agreement (89.6%), reflecting scores concentrated at a single level.

**Table 3. Agreement between Claude-Opus-5.5 and Gemini-3.8-Flash across 240 answers.** Exact agreement is the proportion of answers given the same score by both judges; Cohen's κ measures inter-rater agreement between two LLM judges, adjusted for chance agreement; Pearson r is the correlation between the two judges' scores. n is the number of answers scored on the module by both judges.

| Module | n | Exact agreement | Cohen's κ | Pearson r |
|---|---|---|---|---|
| A. Numerical accuracy | 240 | 83.6% | 0.832 | 0.845 |
| B. Computational correctness | 98 | 77.3% | 0.812 | 0.817 |
| C. Evidential grounding | 240 | 80.4% | 0.721 | 0.736 |
| D. Medical safety | 240 | 89.6% | 0.472 | 0.577 |
| E. Consistency & context | 240 | 86.3% | 0.718 | 0.723 |
| F. Usefulness | 124 | 68.5% | 0.464 | 0.538 |
| G. Compound-query handling | 240 | 87.9% | 0.734 | 0.751 |
| **Weighted total** | **240** | - | - | **0.878** |

Critical failures occurred at similar rates in the two configurations (18/120 vs. 16/120 answers), and tag instances were also similar (22 vs. 20; Table B2). Multi Agent had fewer numerical hallucinations (6 vs. 2) and missing-evidence tags (5 vs. 3) but more status misclassifications (3 vs. 7), trend-direction errors (0 vs. 4), and statements naming or dosing medications (0 vs. 2).

## 4.5. Effect by Query Type and Agent Group

We explored two post hoc subgroup definitions: the presence of a judgment intent (Table 4) and the required agent group (Table 5). In the subset of 25 queries containing a judgment intent (numeric, categorical, or free-text judgment of a user-supplied value), Multi Agent increased the weighted score by 0.190 points ($p = 0.006$; Table 4). The remaining 95 queries showed a smaller improvement of 0.079 points, which was not statistically significant ($p = 0.153$).

**Table 4. Effect split defined by the presence of a judgment intent.**

| Subset | n | Single Agent | Multi Agent | Δ | p |
|---|---|---|---|---|---|
| Judgment queries (user-supplied value) | 25 | 1.733 | 1.923 | +0.190 | **0.006** |
| All other queries | 95 | 1.685 | 1.764 | +0.079 | 0.153 |

The agent-group analysis showed a related pattern (Table 5). The improvement was significant for queries requiring record lookup ($\Delta = +0.169$, $p < 0.001$) or numeric judgment ($\Delta = +0.174$, $p = 0.026$), particularly when record lookup was combined with cross-year comparison ($\Delta = +0.279$, $p = 0.022$), medical knowledge ($\Delta = +0.198$, $p = 0.003$), or numeric judgment ($\Delta = +0.186$, $p = 0.048$). Queries requiring medical knowledge, supplement recommendation, cross-year comparison, or result interpretation did not show statistically significant differences. Because the agent groups overlap, these subgroup analyses should be interpreted as exploratory rather than as independent comparisons. Overall, the results suggest that the benefit of Multi Agent execution is concentrated in compound queries that combine personal-record lookup with other requirements, rather than uniformly across compound queries.

**Table 5. Weighted total scores by required agent group (Gemini-3.8-Flash).** Agent and pairwise rows may overlap because queries can require multiple agents. Rows by number of required agents partition 119 queries; one remaining query requires two record lookups handled by the same agent. Agents and agent pairs required by fewer than ten queries are omitted. Δ denotes Multi Agent minus Single Agent.

| Required agent(s) | n | Single Agent | Multi Agent | Δ | p |
|---|---|---|---|---|---|
| Numeric judgment | 18 | 1.719 | 1.893 | +0.174 | **0.026** |
| Record lookup | 86 | 1.718 | 1.887 | +0.169 | **< 0.001** |
| Supplement recommendation | 24 | 1.655 | 1.755 | +0.100 | 0.321 |
| Result interpretation | 22 | 1.589 | 1.665 | +0.076 | 0.567 |
| Medical knowledge | 71 | 1.728 | 1.795 | +0.067 | 0.306 |
| Cross-year comparison | 41 | 1.594 | 1.621 | +0.027 | 0.769 |
| Cross-year comparison + Record lookup | 13 | 1.523 | 1.803 | +0.279 | **0.022** |
| Record lookup + Result interpretation | 10 | 1.500 | 1.723 | +0.223 | 0.280 |
| Record lookup + Medical knowledge | 44 | 1.746 | 1.944 | +0.198 | **0.003** |
| Record lookup + Numeric judgment | 14 | 1.702 | 1.889 | +0.186 | **0.048** |
| Record lookup + Supplement recommendation | 17 | 1.786 | 1.816 | +0.030 | 0.812 |
| Cross-year comparison + Medical knowledge | 26 | 1.664 | 1.573 | −0.090 | 0.488 |
| Medical knowledge + Result interpretation | 12 | 1.819 | 1.689 | −0.130 | 0.404 |
| 2 required agents | 87 | 1.723 | 1.815 | +0.092 | 0.089 |
| 3 required agents | 26 | 1.647 | 1.758 | +0.112 | 0.308 |
| 4 required agents | 6 | 1.445 | 1.706 | +0.260 | 0.160 |

## 4.6. The Cost of Multi Agent Execution

Multi Agent increased latency, model usage, and response length (Table 6). Mean time to first token increased from 13.22 to 17.04 s (1.29×), and total latency from 14.72 to 19.22 s (1.31×). Mean input tokens, output tokens, and LLM calls increased by 1.56×, 1.90×, and 1.70×, respectively, while cost per query increased from USD 0.0095 to USD 0.0193 (2.02×). These increases reflect the additional agent executions, the per-agent extraction calls of record-based agents, and the larger synthesis input. Although LLM calls increased by 1.70×, total latency increased by only 1.31×, reflecting concurrent agent execution. Multi Agent answers were also longer than Single Agent answers, increasing from 435 to 583 characters (1.34×). Because longer responses may cover more requirements, response length may contribute to the observed quality difference; however, the present analysis does not isolate this effect from the number of executed agents. In the human pairwise evaluation (Section 4.7), among the 22 queries whose answers differed in length by no more than 15%, the evaluators preferred the Multi Agent answer in 59.1% (13/22; $p = 0.52$) and 50.0% (11/22; $p = 1.00$) of cases, so a contribution of response length to human preference cannot be ruled out.

**Table 6. Efficiency and response length on the test set.** Values are paired means ± standard deviations. Ratio denotes Multi Agent divided by Single Agent. P-values are from paired t-tests.

| Metric | Single Agent | Multi Agent | Ratio | p |
|---|---|---|---|---|
| Time to first token (s) | 13.22 ± 5.28 | 17.04 ± 4.66 | 1.29× | **< 0.001** |
| Total latency (s) | 14.72 ± 5.76 | 19.22 ± 5.15 | 1.31× | **< 0.001** |
| LLM calls | 2.98 ± 1.01 | 5.08 ± 1.03 | 1.70× | **< 0.001** |
| Input tokens | 17,580 ± 5,619 | 27,470 ± 5,002 | 1.56× | **< 0.001** |
| Output tokens | 809 ± 325 | 1,535 ± 614 | 1.90× | **< 0.001** |
| Cost (USD/query) | 0.0095 ± 0.0045 | 0.0193 ± 0.0053 | 2.02× | **< 0.001** |
| Answer length (chars) | 435 ± 221 | 583 ± 257 | 1.34× | **< 0.001** |

## 4.7. Human Evaluation

Two evaluators who were not authors of this study conducted the human evaluation. Both were quality assurance specialists at DaNaA Data Co., Ltd. responsible for verifying chatbot responses, had no clinical training, and were not involved in developing the evaluated system. Before the evaluation, they received a one-page guideline outlining the comparison criteria but were not informed of the system configurations. Each evaluator independently compared the Single Agent and Multi Agent answers for the same 120 queries in a blinded forced-choice evaluation. The two answers were randomly assigned to presentation slots A and B, and the evaluators did not discuss their judgments. Both evaluators preferred the Multi Agent answer in most comparisons (66.7% and 68.3%; $p < 0.001$; Table 7) and agreed on 110 of 120 queries (Cohen's $\kappa = 0.810$).

**Table 7. Human pairwise preference under blinded randomized presentation.** Preference denotes the proportion favoring Multi Agent. The last row reports inter-rater agreement as Cohen's κ with an asymptotic 95% CI.

| Evaluation | n | Wins (Multi Agent / Single Agent) | Multi Agent preference (95% CI) | p |
|---|---|---|---|---|
| Evaluator 1 | 120 | 80 / 40 | 66.7 (57.8-74.5) | **<0.001** |
| Evaluator 2 | 120 | 82 / 38 | 68.3 (59.6-76.0) | **<0.001** |
| Queries with rater agreement | 110 | 76 / 34 | 69.1 (59.9-77.0) | **<0.001** |
| Cohen's κ (95% CI) | 120 | 0.810 (0.697-0.923) | | |

Table 8 compares each LLM judge with the two evaluators at the query level. Judge preference was taken as the answer with the higher weighted total. Because forced-choice evaluation does not allow ties, for queries whose two answers received the same total score (19, 17, 46, and 28 queries for Claude-Opus-5.5, GPT-6-Sol, Gemini-3.8-Flash, and Solar-Pro-4, respectively), a separate forced-choice evaluation was conducted in which the same judge was given the same rubric and reference records and asked to select the relatively better of the two answers. For Solar-Pro-4, 12 of the 28 tied queries had no forced-choice result, so its agreement is computed over the remaining 108 queries. The judges agreed with the evaluators on 90.0% (Claude-Opus-5.5), 78.3–80.0% (GPT-6-Sol), 84.2–89.2% (Gemini-3.8-Flash), and 73.1–75.0% (Solar-Pro-4) of queries. Every judge–evaluator pair agreed less than the two evaluators did with each other (Cohen's κ = 0.363–0.778 vs. 0.810), indicating that although LLM judges show a certain level of agreement with human evaluators, their agreement falls short of inter-rater agreement, and they may therefore be better used to complement human evaluation than to fully replace it.

The judges showed no systematic direction relative to the evaluators: they preferred Multi Agent in 65.0–72.2% of queries, compared with 66.7% and 68.3% for the evaluators. Preferences set by the forced-choice re-query agreed with the evaluators less often than score-based preferences for GPT-6-Sol (58.8–64.7% vs. 81.6–82.5%) and Gemini-3.8-Flash (76.1–87.0% vs. 89.2–90.5%). Because evaluators compared answer pairs whereas judges mostly scored answers individually, these results reflect how well judge scores reproduce human preference rather than direct agreement on a shared task.

**Table 8. Agreement between LLM judges and human evaluators.** Agreement is the proportion of queries on which the judge's preference matched the evaluator's choice (n = 120 for each evaluator and 110 for queries with rater agreement; 108 and 100 for Solar-Pro-4). Judge preference is the answer with the higher weighted total; ties were resolved by a forced-choice re-query to the same judge. The last row reports Cohen's κ against Evaluator 1 and Evaluator 2.

| Agreement | Claude-Opus-5.5 | GPT-6-Sol | Gemini-3.8-Flash | Solar-Pro-4 |
|---|---|---|---|---|
| Evaluator 1 | 90.0 (CI: 83.3-94.2) | 80.0 (CI: 72.0-86.2) | 84.2 (CI: 76.6-89.6) | 73.1 (CI: 64.1-80.6) |
| Evaluator 2 | 90.0 (CI: 83.3-94.2) | 78.3 (CI: 70.1-84.8) | 89.2 (CI: 82.3-93.6) | 75.0 (CI: 66.1-82.2) |
| Queries with rater agreement | 93.6 (CI: 87.4-96.9) | 81.8 (CI: 73.6-87.9) | 90.0 (CI: 83.0-94.3) | 76.0 (CI: 66.8-83.3) |
| Cohen's κ (Evaluator 1/2) | 0.778 / 0.775 | 0.556 / 0.513 | 0.637 / 0.748 | 0.363 / 0.396 |

# 5. Discussion and Limitations

The fixed intent taxonomy and intent-to-agent mapping trade flexibility for control. Each query type has an auditable destination, context scope, and configured model, enabling path-specific grounding and safeguards for judging user-supplied values while allowing stage-specific model selection. However, this structure requires manual maintenance and remains dependent on probabilistic intent classification: new capabilities require updates to the taxonomy and routing map, and deterministic execution begins only after classification.

Several limitations should be considered. The benchmark consists only of compound Korean queries and therefore does not represent the full distribution of real-world traffic. Each configuration generated a single answer per query, so generation variability across repeated runs was not assessed. The system is intended to provide health guidance rather than diagnosis. Although the LLM-based evaluations were complemented by human pairwise evaluation, the human evaluation involved only two non-clinical evaluators and therefore does not establish clinical validity. In addition, the evaluation used five fixed synthetic screening records rather than real patient data and did not assess the effects of persistent memory.

Future work should compare against a variant of the current multi-agent system in which the Multi-Agent Orchestrator is restructured to operate with full autonomy, without intent-classification guidance, investigate learned or feedback-driven routing using execution traces, and validate responses with clinical experts. In addition, these limitations should be addressed by measuring generation variability across repeated runs, extending evaluation to additional screening schemas and languages, and evaluating the effects of memory.

# 6. Conclusion

We evaluated intent-based Multi Agent execution for personalized health-checkup guidance on 120 compound Korean queries. Compared with the Single Agent configuration, Multi Agent increased the weighted LLM-judge score from 1.695 to 1.797, and the direction of improvement was consistent across all four LLM judges from different providers. The gain came from usefulness, consistency, and the handling of every requirement in compound queries, whereas medical safety did not differ, numerical accuracy and grounding improved under only one of the four judges, and critical failures occurred at similar rates. In blinded pairwise comparisons, two human evaluators preferred the Multi Agent answer in 66.7% and 68.3% of queries. Exploratory subgroup analyses suggested that the gain was concentrated in queries involving personal-record lookup, while latency and cost increased by 1.31× and 2.02×, respectively. These results motivate selective Multi Agent execution for queries whose requirements span distinct agent roles, while clinician validation and controlled comparisons remain necessary.

# Ethics Statement

This study utilized entirely synthetic health checkup data generated under clinically reviewed in-house criteria. No actual patient records or personally identifiable information were used in the data generation or evaluation process. Therefore, IRB approval was not required for this study.

# Conflict of Interest

All authors and both evaluators are employees of DaNaA Data Co., Ltd., the developer of the evaluated system. The authors declare no other competing interests.

# References

[1] National Health Insurance Service. 2023 National Health Screening Statistical Yearbook. National Health Insurance Service, Republic of Korea, 2024. In Korean.

[2] D. W. Shin et al. National General Health Screening Program in Korea: History, Current Status, and Future Direction: A Scoping Review. Precision and Future Medicine, 6(1):9–31, 2022.

[3] K. Singhal et al. Large Language Models Encode Clinical Knowledge. Nature, 620(7972):172–180, 2023.

[4] K. Singhal et al. Toward Expert-Level Medical Question Answering with Large Language Models. Nature Medicine, 31(3):943–950, 2025.

[5] J. W. Ayers et al. Comparing Physician and Artificial Intelligence Chatbot Responses to Patient Questions Posted to a Public Social Media Forum. JAMA Internal Medicine, 183(6):589–596, 2023.

[6] N. F. Liu et al. Lost in the Middle: How Language Models Use Long Contexts. Transactions of the Association for Computational Linguistics, 12:157–173, 2024.

[7] A. Pal, L. K. Umapathi, and M. Sankarasubbu. Med-HALT: Medical Domain Hallucination Test for Large Language Models. In Proceedings of the 27th Conference on Computational Natural Language Learning (CoNLL), pages 314–334, 2023.

[8] S. Pandit et al. MedHallu: A Comprehensive Benchmark for Detecting Medical Hallucinations in Large Language Models. In Proceedings of the 2025 Conference on Empirical Methods in Natural Language Processing (EMNLP), pages 2858–2873, 2025.

[9] X. Tang et al. MedAgents: Large Language Models as Collaborators for Zero-Shot Medical Reasoning. In Findings of the Association for Computational Linguistics: ACL 2024, pages 599–621, 2024.

[10] Q. Wu et al. AutoGen: Enabling Next-Gen LLM Applications via Multi-Agent Conversation. In First Conference on Language Modeling (COLM), 2024.

[11] S. Hong et al. MetaGPT: Meta Programming for a Multi-Agent Collaborative Framework. In The Twelfth International Conference on Learning Representations (ICLR), 2024.

[12] G. Li et al. CAMEL: Communicative Agents for "Mind" Exploration of Large Language Model Society. In Advances in Neural Information Processing Systems 36 (NeurIPS), pages 51991–52008, 2023.

[13] T. Guo et al. Large Language Model Based Multi-Agents: A Survey of Progress and Challenges. In Proceedings of the 33rd International Joint Conference on Artificial Intelligence (IJCAI), pages 8048–8057, 2024.

[14] Y. Kim et al. MDAgents: An Adaptive Collaboration of LLMs for Medical Decision-Making. In Advances in Neural Information Processing Systems 37 (NeurIPS), pages 79410–79452, 2024.

[15] L. Chen, M. Zaharia, and J. Zou. FrugalGPT: How to Use Large Language Models While Reducing Cost and Improving Performance. Transactions on Machine Learning Research, 2024.

[16] I. Ong et al. RouteLLM: Learning to Route LLMs from Preference Data. In The Thirteenth International Conference on Learning Representations (ICLR), 2025.

[17] Y. Shen et al. HuggingGPT: Solving AI Tasks with ChatGPT and Its Friends in Hugging Face. In Advances in Neural Information Processing Systems 36 (NeurIPS), pages 38154–38180, 2023.

[18] H. Zhang et al. HuatuoGPT, Towards Taming Language Model to Be a Doctor. In Findings of the Association for Computational Linguistics: EMNLP 2023, pages 10859–10885, 2023.

[19] J. Khasentino et al. A Personal Health Large Language Model for Sleep and Fitness Coaching. Nature Medicine, 31(10):3394–3403, 2025.

[20] S. Kweon et al. KorMedMCQA: Multi-Choice Question Answering Benchmark for Korean Healthcare Professional Licensing Examinations. arXiv preprint arXiv:2403.01469, 2024.

[21] P. Lewis et al. Retrieval-Augmented Generation for Knowledge-Intensive NLP Tasks. In Advances in Neural Information Processing Systems 33 (NeurIPS), pages 9459–9474, 2020.

[22] J. Johnson, M. Douze, and H. Jégou. Billion-Scale Similarity Search with GPUs. IEEE Transactions on Big Data, 7(3):535–547, 2021.

[23] Y. Liu et al. G-Eval: NLG Evaluation Using GPT-4 with Better Human Alignment. In Proceedings of the 2023 Conference on Empirical Methods in Natural Language Processing (EMNLP), pages 2511–2522, 2023.

[24] L. Zheng et al. Judging LLM-as-a-Judge with MT-Bench and Chatbot Arena. In Advances in Neural Information Processing Systems 36 (NeurIPS), Datasets and Benchmarks Track, pages 46595–46623, 2023.

[25] Y. Lee et al. CheckEval: A Reliable LLM-as-a-Judge Framework for Evaluating Text Generation Using Checklists. In Proceedings of the 2025 Conference on Empirical Methods in Natural Language Processing (EMNLP), pages 15771–15798, 2025.

[26] C.-H. Chiang and H.-Y. Lee. A Closer Look into Using Large Language Models for Automatic Evaluation. In Findings of the Association for Computational Linguistics: EMNLP 2023, pages 8928–8942, 2023.

# Appendix A. Intent Taxonomy

The deployed taxonomy contains 29 intent codes, each assigned by hand to one of the seven intent-bearing agent categories (Table A1). The classifier selects codes, but the static map determines the executing agent within the category.

**Table A1. Deployed intent taxonomy grouped by agent category.**

| Category | Code | Intent | Category | Code | Intent |
|---|---|---|---|---|---|
| **Data analysis** | INT104 | Personal result lookup | **Medical knowledge** | INT001 | Lifestyle improvement |
| | INT105 | Status verdict lookup | | INT002 | Test-disease association |
| | INT106 | Overall-opinion lookup | | INT003 | Reference range |
| | INT013 | Whole-report summary | | INT004 | Finding explanation |
| | INT015 | Cross-year comparison | | INT005 | Finding interpretation |
| **Judgment** | INT108 | Numeric judgment | | INT006 | Composite analysis |
| | INT109 | Categorical judgment | | INT007 | Contributing factors |
| | INT110 | Free-text judgment | | INT008 | Test description |
| **Nutrition** | INT111 | Supplement recommendation | | INT009 | Concept comparison |
| **Hospital search** | INT016 | Hospital search | | INT010 | General screening knowledge |
| **Conversation** | INT101 | Small talk / off-domain | | INT011 | Follow-up action |
| | INT198 | Personal-data refusal | | INT102 | General medical question |
| | INT199 | Out-of-scope refusal | **Service** | INT012 | Chatbot usage |
| | INT200 | Injection / system probe | | INT014 | Promotion / event |
| | - | - | | INT103 | Administrative inquiry |

# Appendix B. Evaluation Rubric

The rubric contains seven modules scored from 0 to 2 and combined into a weighted total out of 2.0 (Table B1), with non-applicable weights redistributed. Compound-query handling is derived by rule from pre-defined part-level judgments: each requirement is marked as covered, partial, or missing, regardless of correctness, with an additional judgment on whether the parts are combined without contradiction. Judges determine module applicability under the same written criteria; within compound-query handling, an unaddressed requirement is marked missing and scored zero rather than excluded. Critical failures are recorded separately to prevent disqualifying errors from being averaged away. The design is inspired by checklist-based judging [25] and analyze-then-rate prompting [26], adapted here to include a record-checking step before rating (Appendix C).

## B.1. Modules and weights

Table B1. Rubric modules and weights. Numerical accuracy and medical safety are weighted 1.25 times the other modules; the weights sum to 1.

| Module | Weight | Description |
|---|---|---|
| A. Numerical accuracy | 0.167 | Whether every value and year cited matches the record, and whether normal/abnormal status is stated correctly. |
| B. Computational correctness | 0.133 | Whether differences, rates of change and trend directions are arithmetically right. |
| C. Evidential grounding | 0.133 | Whether each claim is traceable to the screening record or a retrieved document, and whether general statements are kept distinct from the user's own results. |
| D. Medical safety | 0.167 | Whether what the answer states is harmful — a diagnosis asserted, a drug prescribed, an abnormal value called normal, alarm raised over a normal one, an invasive test or emergency care urged for a marginal deviation, personal data or the system prompt exposed. What the answer omits is not scored here. |
| E. Consistency and context | 0.133 | Whether the answer addresses what was asked and whether its form suits the request. |
| F. Usefulness and actionability | 0.133 | Whether the next action is concrete, reflects the user's own context, and handles uncertainty honestly. |
| G. Compound-query handling | 0.133 | Whether each requirement of a compound query is addressed in the form it calls for, and whether the parts are combined without contradiction or fragmentation. |

The module weights are close to uniform (Table B1). Equal weighting yields nearly identical between-system differences (+0.109 for the primary judge and +0.121 for Claude-Opus-5.5, versus +0.103 and +0.111 under the weighted scheme), with a mean absolute difference of 0.009 per answer between weighted and unweighted total scores. Formatting and readability are excluded because they reflect presentation rather than substantive answer quality. Query coverage is scored per requirement in compound-query handling rather than as overall completeness, which may favor verbosity. Medical safety evaluates harmful statements rather than omissions, which are captured by accuracy, grounding, and usefulness. Supplement recommendations are permitted when framed as recommendations, but unsupported efficacy claims are penalized. Hedged or appropriately qualified statements receive 1 rather than 0.

## B.2. Critical-failure gate

The critical-failure gate comprises tags for value and arithmetic errors (numerical hallucination, status misclassification, year mismatch, calculation error, trend direction error, sex-specific range errors, and reversed HDL interpretation), medical safety violations (prescription, asserted diagnosis, unwarranted alarm, false reassurance, over-escalation, and exposure of personal data or the system prompt), grounding and consistency errors (missing evidence, off-topic answer, unsupported medical claim, and within-answer contradiction), and contradictions between parts of a compound answer. A response fails the gate if it contains at least one tag. Tag counts are reported separately because a single response may contain multiple defects (Table B2). The gate therefore captures the presence of a disqualifying error rather than overall answer quality.

**Table B2. Critical-failure tags assigned by the primary judge across the test set**. Tags not observed in either configuration are omitted. Negative Δ indicates fewer failures with Multi Agent. Tag instances may exceed the number of failing answers because an answer can contain multiple tags.

| Failure tag | Single Agent | Multi Agent | Δ | Failure tag | Single Agent | Multi Agent | Δ |
|---|---|---|---|---|---|---|---|
| Status misclassification | 3 | 7 | +4 | Calculation error | 1 | 0 | −1 |
| Missing evidence | 5 | 3 | −2 | Within-answer contradiction | 1 | 0 | −1 |
| Numerical hallucination | 6 | 2 | −4 | Cross-part contradiction | 1 | 0 | −1 |
| Year mismatch | 3 | 2 | −1 | False reassurance | 1 | 0 | −1 |
| Trend direction error | 0 | 4 | +4 | Off-topic answer | 1 | 0 | −1 |
| Prescription provided | 0 | 2 | +2 | **Total (instances)** | **22** | **20** | **−2** |

# Appendix C. Judge Protocol and Independent Validation

Judging follows a fixed sequence: identify applicable modules, fact-check answer values against the corresponding screening records, record the checks, and assign scores, preventing post-hoc justification. Answers are stripped of configuration labels, assigned opaque identifiers, and scored independently. All judges use the same rubric, reference criteria, and screening records under the same blinded protocol. The required response schema includes the six directly scored modules, part-level judgments for compound-query handling, a failure flag and tags, and a brief justification; non-applicable modules are excluded.

Inter-judge variability across the four judges was quantified as a mean per-item standard deviation of 0.168 for weighted total scores and 0.230 for the Multi Agent–Single Agent score differences. Judge-human agreement ranged from 73.1% to 90.0% (Section 4.7). Because the two human evaluators were non-clinicians, this does not establish clinical calibration. The primary judge, Gemini-3.8-Flash, is from a provider not used in the generation pipeline, whereas Claude-Opus-5.5 and GPT-6-Sol share providers with models used in both configurations (Claude-Haiku-4.5 and GPT-5.4-nano). Thus, provider independence holds for Gemini-3.8-Flash and Solar-Pro-4.

**Table C1. Rubric module scores from four LLM judges.** n denotes query pairs in which both answers were scored on the module; it differs across judges because each judge skipped modules it judged not applicable to an answer, most often computational correctness and usefulness.

| Judge model | Module | n | Single Agent | Multi Agent | Δ | p | Sig. |
|---|---|---|---|---|---|---|---|
| Claude-Opus-5.5 | A. Numerical accuracy | 120 | 1.433 ± 0.82 | 1.467 ± 0.77 | +0.033 | 0.725 | |
| | B. Computational correctness | 49 | 1.327 ± 0.80 | 1.388 ± 0.84 | +0.061 | 0.700 | |
| | C. Evidential grounding | 120 | 1.525 ± 0.59 | 1.642 ± 0.53 | +0.117 | 0.071 | |
| | D. Medical safety | 120 | 1.875 ± 0.38 | 1.833 ± 0.46 | −0.042 | 0.426 | |
| | E. Consistency & context | 120 | 1.592 ± 0.51 | 1.733 ± 0.44 | +0.142 | **0.013** | * |
| | F. Usefulness | 62 | 1.210 ± 0.45 | 1.694 ± 0.46 | +0.484 | **<0.001** | *** |
| | G. Compound-query handling | 120 | 1.625 ± 0.57 | 1.842 ± 0.41 | +0.217 | **<0.001** | *** |
| | **Total** | 120 | 1.575 ± 0.42 | 1.686 ± 0.38 | +0.111 | **0.017** | * |
| Gemini-3.8-Flash | A. Numerical accuracy | 120 | 1.542 ± 0.79 | 1.627 ± 0.71 | +0.085 | 0.343 | |
| | B. Computational correctness | 48 | 1.396 ± 0.84 | 1.521 ± 0.82 | +0.125 | 0.452 | |
| | C. Evidential grounding | 120 | 1.650 ± 0.66 | 1.742 ± 0.56 | +0.092 | 0.199 | |
| | D. Medical safety | 120 | 1.975 ± 0.20 | 1.958 ± 0.27 | −0.017 | 0.595 | |
| | E. Consistency & context | 120 | 1.650 ± 0.56 | 1.783 ± 0.43 | +0.133 | **0.018** | * |
| | F. Usefulness | 68 | 1.603 ± 0.55 | 1.868 ± 0.38 | +0.265 | **0.001** | ** |
| | G. Compound-query handling | 120 | 1.717 ± 0.52 | 1.908 ± 0.32 | +0.192 | **<0.001** | *** |
| | **Total** | 120 | 1.695 ± 0.44 | 1.797 ± 0.36 | +0.103 | **0.027** | * |
| GPT-6-Sol | A. Numerical accuracy | 104 | 1.250 ± 0.90 | 1.375 ± 0.87 | +0.125 | 0.247 | |
| | B. Computational correctness | 49 | 1.245 ± 0.85 | 1.306 ± 0.92 | +0.061 | 0.722 | |
| | C. Evidential grounding | 117 | 1.282 ± 0.74 | 1.410 ± 0.70 | +0.128 | 0.144 | |
| | D. Medical safety | 120 | 1.825 ± 0.51 | 1.842 ± 0.52 | +0.017 | 0.801 | |
| | E. Consistency & context | 118 | 1.492 ± 0.55 | 1.619 ± 0.51 | +0.127 | **0.039** | * |
| | F. Usefulness | 54 | 1.241 ± 0.47 | 1.630 ± 0.49 | +0.389 | **<0.001** | *** |
| | G. Compound-query handling | 120 | 1.483 ± 0.62 | 1.742 ± 0.48 | +0.258 | **<0.001** | *** |
| | **Total** | 120 | 1.460 ± 0.47 | 1.602 ± 0.43 | +0.142 | **0.006** | ** |
| Solar-Pro-4 | A. Numerical accuracy | 119 | 1.588 ± 0.80 | 1.798 ± 0.55 | +0.210 | 0.011 | * |
| | B. Computational correctness | 50 | 1.400 ± 0.93 | 1.680 ± 0.68 | +0.280 | 0.080 | |
| | C. Evidential grounding | 119 | 1.622 ± 0.74 | 1.840 ± 0.49 | +0.218 | **0.005** | ** |
| | D. Medical safety | 120 | 2.000 ± 0.00 | 1.992 ± 0.09 | −0.008 | 0.319 | |
| | E. Consistency & context | 120 | 1.617 ± 0.70 | 1.833 ± 0.42 | +0.217 | **0.002** | ** |
| | F. Usefulness | 117 | 1.137 ± 0.37 | 1.462 ± 0.50 | +0.325 | **<0.001** | *** |
| | G. Compound-query handling | 120 | 1.633 ± 0.66 | 1.817 ± 0.47 | +0.183 | **0.005** | ** |
| | **Total** | 120 | 1.613 ± 0.48 | 1.799 ± 0.33 | +0.186 | **<0.001** | *** |